\documentclass[10pt,twocolumn,letterpaper]{article}
\usepackage{iccv}
\usepackage{times}
\usepackage{epsfig}
\usepackage{graphicx}
\usepackage{amsmath}
\usepackage{amssymb}

\usepackage{url}            
\usepackage{booktabs}       
\usepackage{amsfonts}       
\usepackage{nicefrac}       
\usepackage{microtype}      

\usepackage{tabu}           
\usepackage{multirow}
\usepackage{subcaption}
\usepackage{bigstrut}
\usepackage{amsmath}
\usepackage{algorithm}
\usepackage{algorithmicx}
\usepackage{algpseudocode}

\newcommand{\paragrapha}[2]{\vspace{#1}\noindent\textbf{#2}}

\usepackage[breaklinks=true,bookmarks=false,colorlinks]{hyperref}

\iccvfinalcopy 

\def\iccvPaperID{7385} 

\ificcvfinal\fi

\begin{document}

\title{Counterfactual Attention Learning for \\ Fine-Grained Visual Categorization and Re-identification}

\newcommand*\samethanks[1][\value{footnote}]{\footnotemark[#1]}

\author{%
Yongming Rao\thanks{Equal contribution. ~ \textsuperscript{\dag}Corresponding author.}, ~Guangyi Chen\samethanks, ~Jiwen Lu\textsuperscript{\dag}, ~Jie Zhou\\
{Department of Automation, Tsinghua University, China}\\
{State Key Lab of Intelligent Technologies and Systems, China}\\
{Beijing National Research Center for Information Science and Technology, China}\\
{\tt\small \{raoyongming95,guangyichen1994\}@gmail.com; \{lujiwen,jzhou\}@tsinghua.edu.cn}\\
}

\maketitle
\ificcvfinal\thispagestyle{empty}\fi

\begin{abstract}
   Attention mechanism has demonstrated great potential in fine-grained visual recognition tasks. In this paper, we present a counterfactual attention learning method to learn more effective attention based on causal inference. Unlike most existing methods that learn visual attention based on conventional likelihood, we propose to learn the attention with counterfactual causality, which provides a tool to measure the attention quality and a powerful supervisory signal to guide the learning process. Specifically, we analyze the effect of the learned visual attention on network prediction through counterfactual intervention and maximize the effect to encourage the network to learn more useful attention for fine-grained image recognition. Empirically, we evaluate our method on a wide range of fine-grained recognition tasks where attention plays a crucial role, including fine-grained image categorization, person re-identification, and vehicle re-identification. The consistent improvement on all benchmarks demonstrates the effectiveness of our method. Code is available at \url{https://github.com/raoyongming/CAL}.
\end{abstract}

\section{Introduction}
Attention is one of the most fundamental mechanism of human visual perception. When facing a complex scene, humans are able to select regions of interest, and employ attention to narrow down the search and speed up recognition. Many efforts~\cite{wang2017residual, zheng2017learning, sun2018multi, hu2018relation, rao2019learning, gu2018learning, bottom-up, stal, lu2016hierarchical} have been made to model the mechanism of human attention in computer vision systems, which aim to facilitate high-performance recognition by discovering discriminative regions and mitigating the negative effects brought by diverse visual appearance, cluttered backgrounds, occlusions, pose variations, \textit{etc}. Since subtle differences are key to distinguish subordinate visual categories, visual attention mechanism has proven to be especially effective in fine-grained visual recognition tasks and become a core component in many state-of-the-art methods~\cite{sermanet2014attention, liu2016fully, A3d, zheng2017learning, sun2018multi, hu2019see, liu2017end, lan2017deep, xu2018attention,Rao_2017_ICCV}.

\begin{figure}[t]
  \centering
  \includegraphics[width=0.95\linewidth]{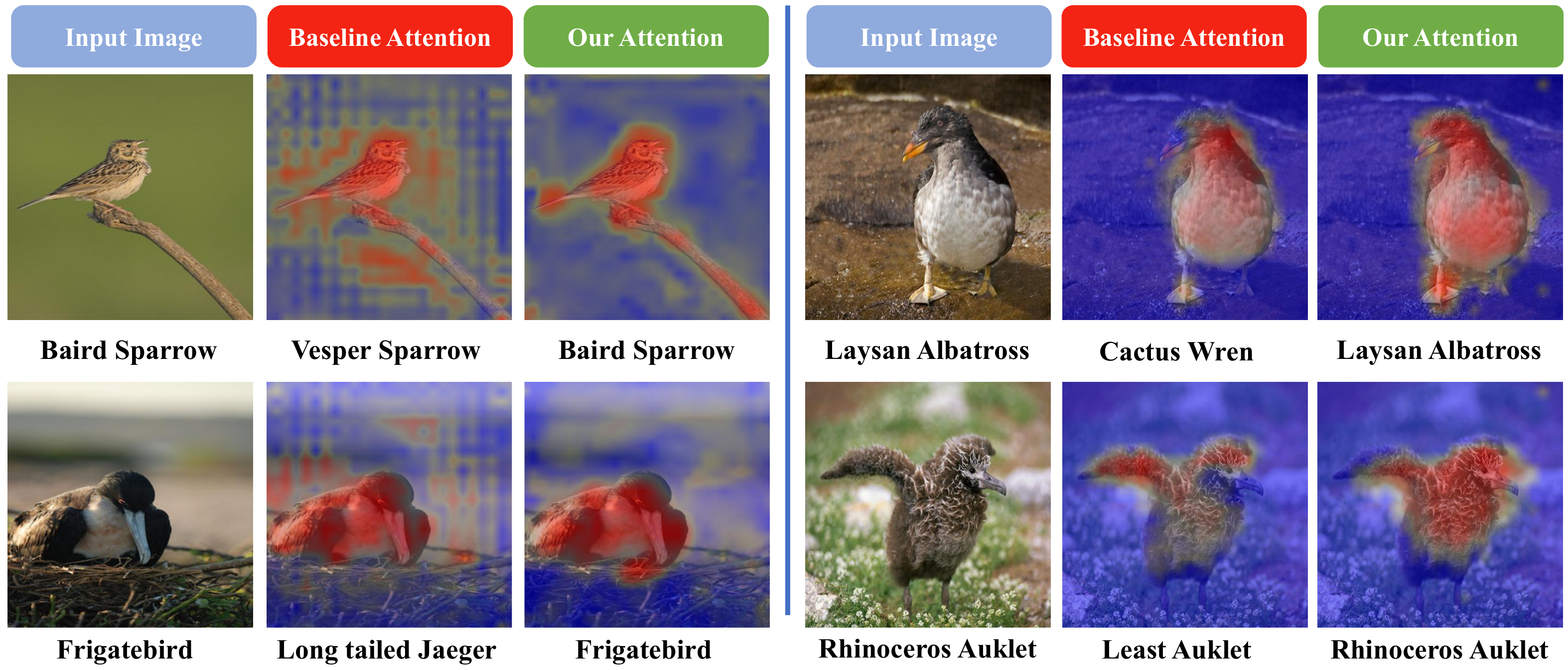}
  \caption{\small Attention visualization on CUB. We respectively show the original images, baseline attention maps, and attention maps with counterfactual learning. In the left part, we observe that our attention maps can better focus on the object. While comparisons in the right part show that our models prefer to look at the whole object instead of some parts. Best in color.
}
\label{fig1}
\vspace{-0.3cm}
\end{figure}

\begin{figure}[t]
  \centering
  \includegraphics[width=0.98\linewidth]{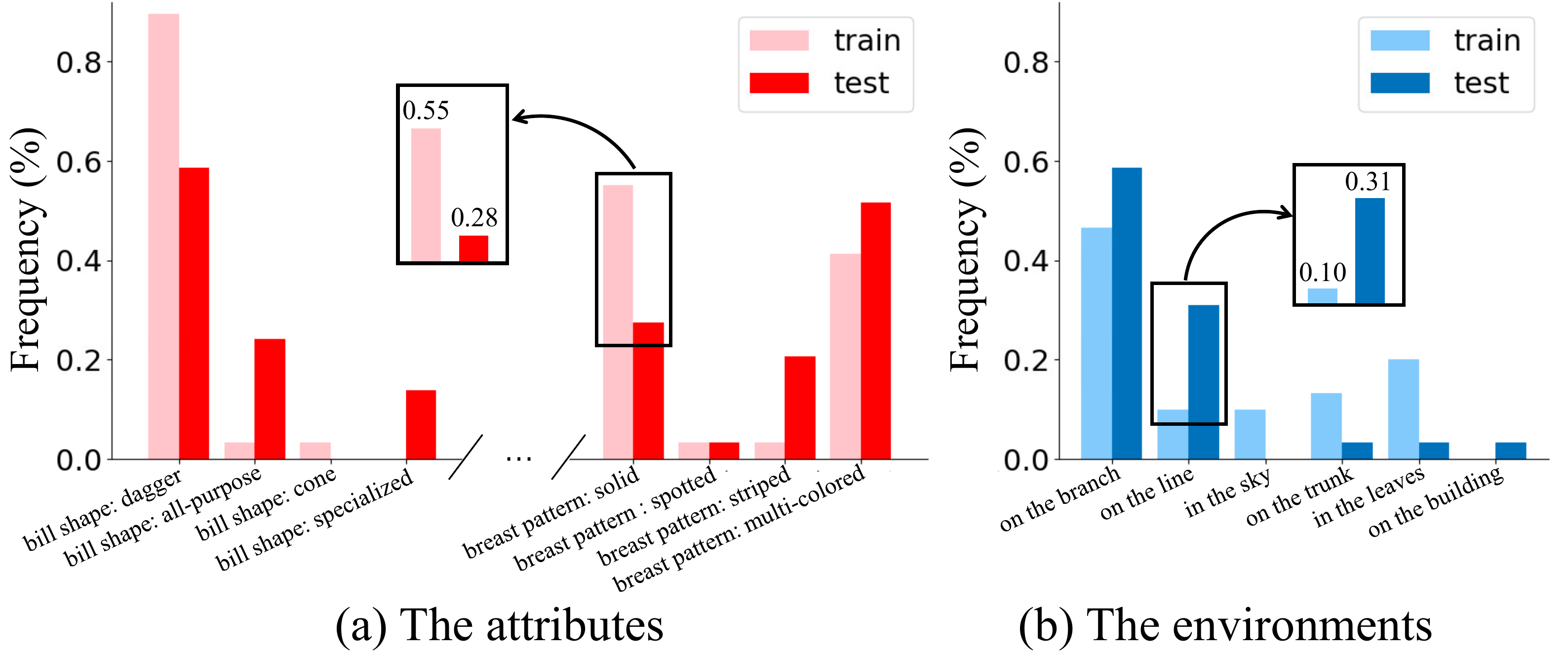}
  \caption{\small The biases of both intrinsic attributes and external environments on CUB. We demonstrate the biases on the training and testing sets by the statistics the frequencies in different attributes and environments, taking the \emph{Ringed Kingfisher} as an example.
}
\label{fig1-2}
\vspace{-0.3cm}
\end{figure}

Despite the widespread use, the problem of how to learn effective attention is still barely studied. Most existing methods learn the visual attention in a weakly-supervised manner, \textit{i.e.}, the attention modules are simply supervised by the final loss function,   without a powerful supervisory signal to guide the training process. This likelihood based approach only explicitly supervises the final prediction (\emph{e.g.}, class probabilities  for classification task) but ignores the causality between the prediction and attention. Previous methods also did not teach the machine how to distinguish between the main clues and biased clues. For example, if most training samples of one specific class appear with sky as background, then the attention model may be very likely to treat the sky as a discriminative region. Although these biased clues may also be beneficial to the classification on the current datasets, the attention model should only focus on the discriminative patterns, \emph{i.e.} the main clues. Besides, directly learning from data may encourage the model to only focus on some certain attributes of the objects instead of all attributes, which may limit the generalization ability on test set. Therefore, we argue that this attention learning scheme is sub-optimal, where the effectiveness of the learned attentions is not always guaranteed, and the attention may lack discriminative power, clear meaning and robustness. As shown in Figure~\ref{fig1}, misleading and scattered attentions can still be observed from a well-trained attention model and potentially lead to the wrong predictions. To better understand this phenomenon, we analyze the statistics of both intrinsic attributes and external environments on the CUB dataset (see Figure~\ref{fig1-2}), where we use the attributes provided by the dataset and manually collect the environment statistics. We see there are biases for both attributes and environment, which indicates either background and single part are not reliable clues for classification. Therefore, it is desired to design new attention learning method beyond conventional likelihood maximization to mitigate the effects of data biases.     

Because of the lack of effective tool to evaluate the quality of attentions quantitatively, correcting misleading attentions is a very challenging task. One straightforward solution is to use extra annotations like bounding boxes or segmentation masks to obtain the regions of interest explicitly such as~\cite{bottom-up}. However, this kind of method requires considerable cost of human labor and is hard to scale up.  Considering the critical role that attention plays in fine-grained visual recognition tasks, it is necessary to design a method to measure the quality of attentions without additional human supervision and further optimize the learned visual attentions. 

In this paper, we present a \emph{counterfactual attention learning} (CAL) method to enhance attention learning based on causal inference. Specifically, we design a tool to analyze the effects of learned visual attention with counterfactual causality. The basic idea is to quantitate the quality of attentions by comparing the effects of facts (\textit{i.e.}, the learned attentions) and the counterfactuals  (\textit{i.e.}, uncorrected attentions) on the final prediction (\textit{i.e.}, the classification score). Then, we propose to maximize the difference (\emph{i.e.}, \emph{effect} in causal inference literature~\cite{pearl2013direct, vanderweele2015explanation}) to encourage the network to learn more effective visual attentions and reduce the effects of biased training set.

The proposed method is model-agonistic and thus can serve as a plug-and-play module to improve a wide range of visual attention models. Our method is also computational efficient, which only introduces a little extra computation cost during training and brings no computation during inference while can significantly improve attention models. We evaluate our method on three fine-grained visual recognition tasks including fine-grained image categorization (CUB200-2011~\cite{cub}, Stanford Cars~\cite{cars} and FGVC Aircraft~\cite{aircraft}), person re-identification (Market1501~\cite{market1501}, DukeMTMC-ReID~\cite{duke} and MSMT17~\cite{MSMT17}) and vehicle re-identification (Veri-776~\cite{veri776} and VehicleID~\cite{VehicleID}). By applying our method to a multi-head attention baseline model, we demonstrate our method significantly improves the baseline and achieve state-of-the-art results on all benchmarks. 
 
\section{Related Work}

\begin{figure*}
  \centering
  \includegraphics[width=0.82\linewidth]{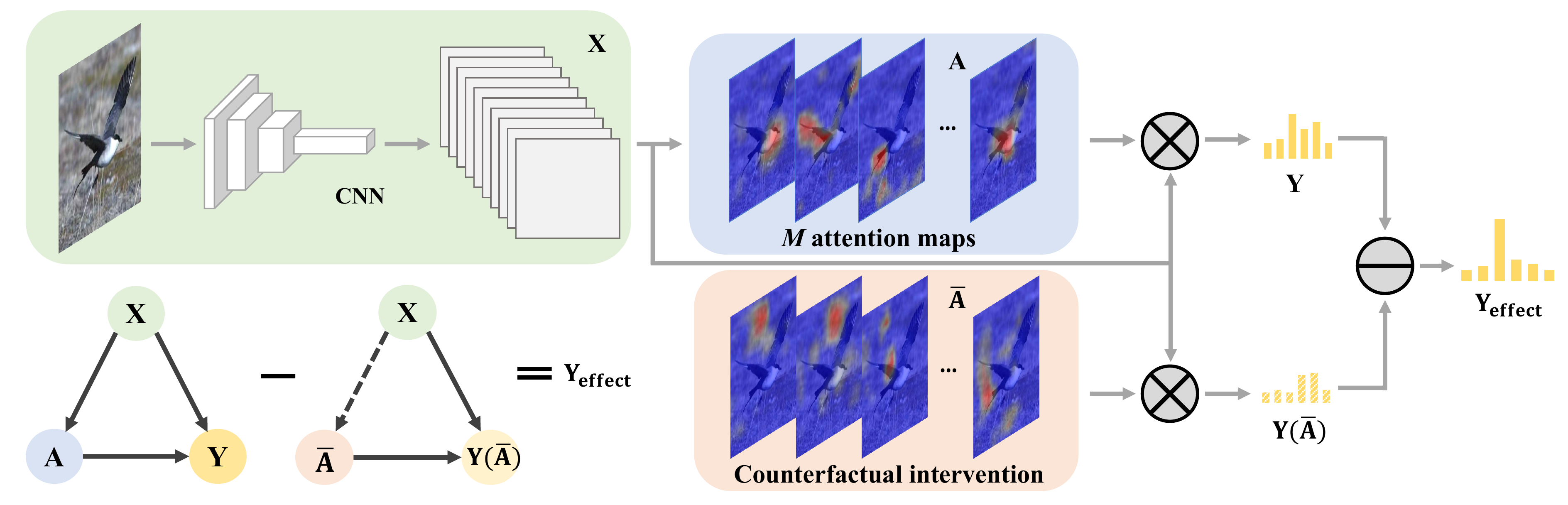}
  \caption{The overall framework of our CAL method.  We first apply the counterfactual intervention for original attention by replacing with random attentions. Then, we subtract the counterfactual classification results from original classification to analyze the effects of learned visual attention and maximize them in the training process.}
\label{fig2}
\vspace{-0.3cm}
\end{figure*}

\paragrapha{0pt}{Fine-Grained Visual Recognition.} Attention mechanism plays an irreplaceable role in fine-grained visual recognition tasks. For example, in fine-grained image categorization task, Sermanet~\textit{et al.}~\cite{sermanet2014attention} pioneer adopting attention mechanism in fine-grained recognition problem and propose a RNN model to learn visual attention. Liu~\textit{et al.}~\cite{liu2016fully} extend the idea and employ a reinforcement learning scheme to obtain visual attentions. The subsequent studies such as MA-CNN~\cite{zheng2017learning}, MAMC~\cite{sun2018multi} and WS-DAN~\cite{hu2019see} further improve this line of methods and design attention models in a bottom-up manner, which achieve very promising results on fine-grained recognition benchmarks. Attention models have also proven to be effective in person/vehicle re-identification problem to handle the image matching misalignment challenge and improve the discriminative power of CNN features.  For instance, Liu~\emph{et al.}~\cite{liu2017end} and Lan~\emph{et al.}~\cite{lan2017deep} employ the attention models to locate the discriminative salient regions in images to improve person re-identification. Xu~\emph{et al.}~\cite{xu2018attention} and Zhao~\emph{et al.}~\cite{zhao2017deeply} design a body part detector to employ the structure of the human body structure in the attention model. Another group of methods~\cite{Liu_2017_CVPR,si2018dual,li2018diversity,chen2020temporal} adopts attention mechanism on video-based person re-identification task to discover key parts in videos. Khorramshahi~\emph{et al.}~\cite{Khorramshahi_2019_ICCV} propose an adaptive attention model and significantly improve the state-of-the-art of vehicle re-identification task.

\paragrapha{5pt}{Causal Reasoning in Vision.} The interest in combining the idea of deep learning and causal reasoning is growing rapidly in recent years. The tool of causality analysis has been successfully used in several areas, including explainable machine learning~\cite{mothilal2020explaining}, fairness~\cite{kusner2017counterfactual}, natural language processing~\cite{wood2018challenges}, reinforcement learning~\cite{kallus2018confounding} and adversarial learning~\cite{kocaoglu2017causalgan}. Some efforts also used causality as an effective tool to alleviate the effects of dataset bias in vision tasks, including  image classification~\cite{lopez2017discovering}, scene graph generation~\cite{tang2020unbiased} and visual commonsense reasoning~\cite{wang2020visual}. In this work, we study causality in the context of visual attention models, which is a new direction that has not been visited.

\section{Approach}

\subsection{Attention Models for Fine-Grained Recognition}
\label{background}

We begin by reviewing the attention models for fine-grained visual recognition, on which our method is built. Given an image $I$ and the corresponding CNN feature maps $\mathbf{X} = f(I)$ of size $H \times W \times C$, visual spatial attention model $\mathcal{M}$ aims to discover the discriminative regions of the image and improve CNN feature maps $\mathbf{X}$ by explicitly incorporating structural knowledge of objects. Note that although some of previous methods like~\cite{wang2017residual} propose to equip the backbone network with spatial attention modules, here we follow the mainstreams~\cite{sermanet2014attention, liu2016fully, zheng2017learning, sun2018multi, hu2019see, liu2017end, liu2017end, lan2017deep, xu2018attention, zhao2017deeply} that learn basic feature maps and attentions separately. Previous studies have demonstrated that this design is more flexible and generic thanks to its model-agnostic nature. 

There have been quite a few variants of $\mathcal{M}$, and we can roughly categorize them into two groups. The first type aims to learn ``hard'' attention maps, where each attention can be represented as a bounding box or segmentation mask that covers a certain region of interest. This group of methods is usually closely related to object detection and semantic segmentation methods. Examples include recurrent visual attention model~\cite{mnih2014recurrent} and fully convolutional attention network~\cite{liu2016fully}. Different from hard-attention models, a wider range of attention models are based on learning ``soft'' attention maps, which are more easy to optimize. In this paper, we focus on studying this group of methods. Specifically, our baseline model adopts the multi-head attention module used in~\cite{zheng2017learning, sun2018multi, hu2019see}. The attention model is designed to learn the spatial distributions of object's parts, which can be represented as attention maps $\mathbf{A} \in \mathbb{R_+}^{H \times W \times M}$, where $M$ is the number of attentions. Using the attention model $\mathcal{M}$, attention maps can be computed by:
\begin{equation}
    \mathbf{A} \!\!=\!\! \{\mathbf{A}_1, \mathbf{A}_2, ..., \mathbf{A}_M\} \!\!=\!\! \mathcal{M}(\mathbf{X}),
\end{equation}
where $\mathbf{A}_i \in \mathbb{R_+}^{H \times W}$ is the attention map covering a certain part, such as the wing of a bird or the cloth of a person. The attention model $\mathcal{M}$ is implemented using a 2D convolutional layer followed by ReLU activation. The attention maps then are used to softly weight the feature maps and aggregate by global average pooling operation $\varphi$: 
\begin{equation} \label{eq2} 
\mathbf{h}_i \!\!=\!\! \varphi(\mathbf{X} * \mathbf{A}_i) \!\!=\!\! \frac{1}{HW} \sum_{h=1}^{H}\sum_{w=1}^{W} \mathbf{X}^{h,w}\mathbf{A}_i^{h,w},   
\end{equation}
where $*$ denotes element-wise multiplication for two tensors. Following the practice in~\cite{hu2019see}, we summarize the representation of different parts to form the global representation $\mathbf{h}$ :
\begin{equation} \label{eq3}
\mathbf{h} \!\!=\!\! \texttt{normalize}([\mathbf{h}_1, \mathbf{h}_2, ..., \mathbf{h}_M]),   
\end{equation}
where we concatenate these representations and normalize the summarized representation to control its scale. The final representation $\mathbf{h}$ can be fed into a classifier (\emph{e.g.}, fully connected layer) for image classification task or a distance metric (\emph{e.g.}, Euclidean distance) for image retrieval task. The overall framework of our baseline attention model is illustrated in Figure~\ref{fig2}.

\subsection{Attention Models in Causal Graph}
\label{graph}

Before we show our counterfactual method, we first introduce how to reformulate the above model in the language of causal graph.  Causal graph is also known as structural causal model, which is a directed acyclic graph $\mathcal{G} \!\!=\!\! \{\mathcal{N}, \mathcal{E}\}$. Each variable in the model has a corresponding node in $\mathcal{N}$ while the causal links $\mathcal{E}$ describe how these variable interact with each other. As presented in Figure~\ref{fig2}, we can use nodes in the causal graph to represent variables in the attention model, including the CNN feature maps (or the input image) $X$, the learned attention maps $A$ and the final prediction $Y$. The link $X \to A$ represents that the attention model takes as input the CNN feature maps and produces corresponding attention maps. $(X, A) \to Y$ indicates the feature maps and attention maps jointly determine the final prediction. Causal relations between nodes are encoded in the links $\mathcal{E}$, where we call node $X$ is the causal parent of $A$, and $Y$ is the causal child of $X$ and $A$. Note that since we do not impose any constraints on the network architecture of backbone models and the implementation details of the attention model, the causal graph can also represent many other attention models. Therefore, our method is model-agonistic and thus can also be extended to a wider range of attention learning problems.

\subsection{Counterfactual Attention Learning}
\label{learning}

Conventional likelihood methods optimize the attention by only supervising the final prediction $Y$ and regard the model as a black box, which ignores how the learned attention maps affect the prediction. On the contrary, causal inference~\cite{pearl2018book} provides a tool to help us think out of the black box by analyzing the causalities between variables. Therefore, we propose to employ the causalities to measure the quality of the learned attention and then improve the model by encourage the network to produce more influential attention maps. 

By introducing the causal graph, we can analyze causalities by directly manipulate the values of several variable and see the effect. Formally, the operation is termed \emph{intervention} in causal inference literature, which can be denoted as $do(\cdot)$. When we want to investigate the effect of a variable, the intervention operation is performed by wiping out all the in-coming links of the variable and assigning a certain value to the variable. For example, $do(A\!\!=\!\!\bar{\mathbf{A}})$ in our causal graph means we demand the variable $A$ to take the value $\bar{\mathbf{A}}$ and cut-off the link $X \to A$ to force the variable to no longer be caused by its causal parent $X$. 

Inspired by causal inference methods~\cite{pearl2013direct, vanderweele2015explanation}, we propose to adopt \emph{counterfactual intervention} to investigate the effects of the learned visual attention. The counterfactual intervention is achieved by an imaginary intervention altering the state of the variables assumed to be different~\cite{vanderweele2015explanation, hagmayer2007causal}. In our case, we conduct counterfactual intervention $do(A\!\!=\!\!\bar{\mathbf{A}})$ by imagining non-existent attention maps $\bar{\mathbf{A}}$ to replace the learned attention maps and keeping the feature maps $X$ unchanged. We can obtain the final prediction $Y$ after the intervention $A\!\!=\!\!\bar{\mathbf{A}}$ according to (\ref{eq2}) and~(\ref{eq3}):
\begin{equation}
 Y(do(A\!\!=\!\!\bar{\mathbf{A}}), X\!\!=\!\!\mathbf{X}) \!\!=\!\!  \mathcal{C}([\varphi(\mathbf{X}\!\!*\!\!\bar{\mathbf{A}}_1), ..., \varphi(\mathbf{X}\!\!*\!\!\bar{\mathbf{A}}_M)]),
 \end{equation}
 where $\mathcal{C}$ is the classifier. In practice, we can use random attention, uniform attention or reversed attention as the counterfactuals. Evaluation on these options can be found in Section~\ref{sec:ablation}. 
 
 Following~\cite{pearl2013direct, vanderweele2015explanation, tang2020unbiased}, the actual effect of the learned attention on the prediction can be represented by the difference between the observed prediction $Y(A=\mathbf{A}, X=\mathbf{X})$ and its counterfactual alternative $Y(do(A=\bar{\mathbf{A}}), X=\mathbf{X})$:
  \begin{equation} \label{eq_define}
 Y_\text{effect}\!\!=\!\! \mathbb{E}_{\bar{\mathbf{A}} \sim \gamma} [ Y(A\!\!=\!\!\mathbf{A}, X\!\!=\!\!\mathbf{X}) \!\!-\!\! Y(do(A\!\!=\!\!\bar{\mathbf{A}}), X\!\!=\!\!\mathbf{X})],
 \end{equation}
 where we denote the effect on the prediction as $Y_\text{effect}$ and $\gamma$ is the distribution of counterfactual attentions. Intuitively, the effectiveness of an attention can be interpreted as how the attention improves the final prediction compared to wrong attentions. Therefore, we can use $Y_\text{effect}$ to measure the quality of a learned attention. 
 
 Furthermore, we can use the metric of attention quality as a supervision signal to explicitly guide the attention learning process. The new objective can be formulated as:
   \begin{equation}
\mathcal{L} \!\!=\!\! \mathcal{L}_{ce}(Y_\text{effect}, y) \!\!+\!\!  \mathcal{L}_\text{others},
 \end{equation}
where $y$ is the classification label,  $\mathcal{L}_{ce}$ is the cross-entropy loss, and $ \mathcal{L}_\text{others}$ represents the original objective such as standard classification loss. By optimizing the new objective, what we expect to achieve is two-fold: 1) the  attention model should improve the prediction based on wrong attentions as much as possible, which encourages the attention to discover the most discriminative regions and avoid sub-optimal results; 2) we penalize the prediction based on wrong attentions, which forces the classifier to make decision based more on the main clues instead of the biased clues and reduces the influence of biased training set. 

Note that in practice, it is not necessary to compute the expectation in Equation~(\ref{eq_define}) and we only sample a counterfactual attention for each observed attention during training, which is also consistent with the idea of stochastic gradient descent. Therefore, the extra computational cost introduced by our method is an additional forward of the attention model and the classifier, which is very lightweight compared with the CNN backbone. Besides, our method introduces no additional computation during inference.

\section{Experiments}

We assess the effectiveness of our proposed counterfactual attention learning method on several fine-grained visual recognition tasks including fine-grained image categorization, person re-identification and vehicle re-identification. We take the conventional spatial attention as the baseline and compare our counterfactual attention learning method with the baseline method and other state-of-the-art methods. The experimental settings, implementation details and results for different tasks are described below.

\subsection{Fine-grained Image Categorization}

Fine-grained visual categorization focuses on classifying the subordinate-level classes under a fixed basic-level category, such as species of bird, types of car and types of aircraft. The objects under the same basic-level category are always high structured and with low inter-class variances. Thus, attention is effective to look for the key difference in  detail and discover the discriminative regions. 

\vspace{5pt} \noindent \textbf{Datasets and Experimental Settings. } We conducted experiments on widely used CUB200-2011~\cite{cub}, Stanford Cars~\cite{cars} and  Aircraft~\cite{aircraft} datasets for fine-grained bird, car and aircraft classification. CUB200-2011 is composed of 5,994 training images and 5,794 testing images from 200 species of birds.  Stanford Cars contains 16,185 images of cars from 196 different types and among all collected, 8,144 images are used for training and 8,041 images for testing. FGVC Aircraft consists of 10,000 images of 100 fine-grained aircraft types. Following previous methods, we use 2/3 images for training and 1/3 images for evaluation. 

\vspace{5pt} \noindent \textbf{Implementation Details. }  We adopted the standard ResNet-101~\cite{he2016deep} as the backbone network. For attention model, we set the number of attentions to 32 and use the weakly supervised data augmentation method as suggested by~\cite{hu2019see}. During inference, we use multiple crops and horizontal flipping to boost performance. All experiments are conducted with the same hyper-parameters, including 16 batch size, 448$\times$448 image size, and 1e-5 weight decay. We use 1e-3 initial learning rate and reduce the learning rate by 0.9 times in every 2 epochs.

\vspace{5pt} \noindent \textbf{Results. }  We compared our method with the baseline attention model and the state-of-the-art methods in Table~\ref{tb:fg}. The proposed counterfactual attention method can improve the strong baseline by 1.3\%, 1.5\% and 0.6\% on CUB200-2011, Stanford Cars and Aircraft, respectively. Our method also outperformed previous state-of-the-art methods. Notably, although a stronger backbone (DenseNet-161) is used in recent API-Net~\cite{zhuang2020learning} method, our method can still achieve better performance on all three benchmarks. These results clearly demonstrates the effectiveness of our method.

\begin{table}[!]
\caption{Comparisons of the top-1 classification accuracy (\%) with the SOTA fine-grained image categorization methods on CUB200-2011, Stanford Cars and FGVC Aircraft.}
\label{tb:fg}
\begin{center}
\vspace{-0.3cm}
\renewcommand\tabcolsep{6.0pt}
\begin{tabular}{l|c|c|c}
\hline
\textbf{Method} & \textbf{CUB} & \textbf{Cars} & \textbf{Aircraft} \\
\hline
RA-CNN~\cite{fu2017look} & 85.3 & 92.5 & - \\
MA-CNN~\cite{zheng2017learning} & 86.5 & 92.8 & 89.9 \\
MAMC~\cite{sun2018multi} & 86.5 & 93.0 & - \\
NTS-Net~\cite{yang2018learning} & 87.5 & 93.9 & 91.4 \\
WS-DAN~\cite{hu2019see} & 89.4 &  94.5 &93.0\\
DCL~\cite{chen2019destruction} & 87.8 & 94.5 & 93.0 \\
Stacked LSTM~\cite{ge2019weakly} & 90.4 & - & - \\
API-Net~\cite{zhuang2020learning} & 90.0 & 95.3 & 93.9 \\
 \hline
Baseline & 89.3 & 94.0 & 93.6 \\
Baseline + CAL & \textbf{90.6} & \textbf{95.5} & \textbf{94.2} \\
 \hline
\end{tabular}
\end{center}
\vspace{-0.6cm}
\end{table}

\subsection{Person Re-identification}

\begin{table*}[!]
\caption{Comparisons with the state-of-the-art person ReID methods on the Market1501, DukeMTMC-ReID and  MSMT17.}
\label{comparison_reid}
\begin{center}
\vspace{-0.5cm}
\renewcommand\tabcolsep{8pt}
\begin{tabular}{l|*{3}{c}|*{3}{c}|*{3}{c}}
\hline
\multirow{2}{*}{\bf Method} &\multicolumn{3}{c|}{\bf Market1501} &\multicolumn{3}{c|}{\bf DukeMTMC-ReID} & \multicolumn{3}{c}{\bf MSMT17} \\
\cline{2-10} &{\bf R1} &{\bf R5} &{\bf mAP}& {\bf R1} &{\bf R5} &{\bf mAP}& {\bf R1} &{\bf R5} &{\bf mAP} \\
\hline
HA-CNN~\cite{li2018harmonious}  & 91.2 & - & 75.7  & 80.5 & -& 63.8 &-&-&- \\
Part-aligned~\cite{Suh_2018_ECCV}   & 91.7 & 96.9 & 79.6  & 84.4 & 92.2 & 69.3&-&-&- \\ 
Mancs~\cite{Wang_2018_ECCV}   & 93.1 & - & 82.3  & 84.9 & - & 71.8&-&-&- \\ 
PCB+RPP~\cite{sun2018beyond}  & 93.8 & 97.5 & 81.6  & 83.3 & -& 69.2& 68.2 &-& 40.4  \\ 
IANet~\cite{hou2019interaction} & 94.4&-& 83.1 & 87.1&-& 73.4 &75.5 &85.5& 46.8\\
JDGL~\cite{zheng2019joint} &94.8 &-& 86.0&  86.6& -& 74.8 &77.2&- &52.3 \\
SCAL~\cite{chen2019self}& 95.8 & \textbf{98.7} &89.3 & 88.9 & 95.2 & 79.1 &-&-&- \\
MHN~\cite{Chen_2019_ICCV}& 95.1 & 98.1 & 85.0 & 89.1 & 94.6 & 77.2 &-&-&- \\
SFT~\cite{Luo_2019_ICCV}  & 93.4&-& 82.7 & 86.9&-& 73.2 &  73.6 &- & 47.6\\
OSNet~\cite{Zhou_2019_ICCV} & 94.8 & -& 84.9 &  88.6& - & 73.5 & 78.7 &-& 52.9\\
BAT-Net~\cite{Fang_2019_ICCV} & 95.1 & 98.2 &87.4  & 87.7 & 94.7 & 77.3 &79.5 &89.1 &56.8\\
Auto-ReID~\cite{Quan_2019_ICCV}  &94.5 &-& 85.1 &-&-&- &78.2 &88.2 & 52.5\\
MGN+circleloss~\cite{sun2020circle} & \textbf{96.1} & -&  87.4 &-&-&-&76.9&-& 52.1 \\
 \hline
Baseline & 94.0 & 97.7 & 85.9 & 85.7 & 93.6 & 74.0 & 75.3 & 86.4 &50.5 \\
Baseline + CAL & 94.5 & 97.9 & 87.0 & 87.2 & 94.1 & 76.4 & 79.5 & 89.0 & 56.2 \\
Baseline$^\dagger$  & 94.9 & 98.3 & 89.0 & 88.7 & 94.7 & 78.2 & 81.4 & 90.3 & 59.3 \\
Baseline$^\dagger$ + CAL & 95.5 & 98.5 & \textbf{89.5} & \textbf{90.0} & \textbf{96.1} & \textbf{80.5} &  \textbf{84.2} & \textbf{92.0}& \textbf{64.0} \\
 \hline
\end{tabular}
\end{center}
\vspace{-0.7cm}
\end{table*}

Person re-identification (ReID) is a task to match the query individual from multiple gallery candidates across the non-overlapping camera views. It is a challenging problems because of the intra-class variances due to illumination changes, pose variations, occlusions, and cluttered backgrounds. Attention model has gained great success for person ReID by handling the matching misalignment challenge and enhancing the feature representation~\cite{li2018harmonious,chen2019self,liu2017end}. 

\vspace{5pt} \noindent \textbf{Datasets and Experimental Settings. } We conducted the experiments on three public person re-identification datasets including Market1501~\cite{market1501}, DukeMTMC-reID~\cite{duke} and MSMT17~\cite{MSMT17}. Market1501 consists of 32,668 images of 1,501 identities detected by 6 cameras. The whole dataset is divided into a training set with 12,968 images of 751 identities and a test set containing 3,368 query images and 19,732 gallery images of 750 identities. DukeMTMC-reID dataset is composed of 1,404 persons captured by 8 cameras. Its training set includes 16,522 images of 702 persons, while the test set contains the remaining 702 persons with 2,228 query images and 17,661 gallery images. MSMT17 is one of the largest ReID datasets which contains 4,101 identities and 126,411 images. The training set is composed of 30,248 images of 1,041 identities, while the remaining 3,060 identities are used for testing. 

We followed the settings of ~\cite{li2018harmonious} and \cite{chen2019self} for Market1501 and DukeMTMC-reID datasets, and chose the single-query manner to validate our method. For MSMT17, we followed the settings in the test settings in~\cite{MSMT17} and~\cite{zheng2019joint}. We employed the cumulative matching characteristic (CMC) curve and mean average precision (mAP) as the evaluation metrics.

\vspace{5pt} \noindent \textbf{Implementation Details. } We adopted two basic network structures including a standard ResNet50 backbone and the backbone network in SCAL~\cite{chen2019self} which adds 4 channel attention blocks. We applied the modification backbone due to its competitive performance and relatively concise structure, since recent state-of-the-art methods usually use stronger or more sophisticated backbone such as part model~\cite{sun2018beyond,mgn} to improve performance. We used $\dagger$ to indicate the models using the modified backbone. The baseline attention block is same as the one in fine-grained categorization task and we set $M$ to 8 for re-identification models. All experiments are conducted with the same hyper-parameters including 80 batch size, $384\times192$ image size, and 2e-4 learning rate. The data augmentation methods includes random cropping, erasing and horizontal flipping. We trained the network for 160 epochs with triplet loss and softmax loss with learning rate reducing by 10 times in every 40 epochs.

\vspace{5pt} \noindent \textbf{Results. } As shown in Table~\ref{comparison_reid}, we observed that our CAL methods achieve consistent improvement for different baselines on all benchmarks. Specifically, compared with the strong baseline we obtained 0.6\%/0.5\% Rank-1/mAP improvement on the Market1501 dataset, 1.6\%/2.3\% on the DukeMTMC-ReID dataset, and 2.8\%/4.7\% on the MSMT17 dataset. The improvement on the MSMT17 dataset is larger than other two datasets, since images in MSMT17 have larger intra-class variances. Besides, with our strong attention model, we can achieve the SOTA performance on DukeMTMC-ReID and MSMT17 datasets.

\subsection{Vehicle Re-identification}

Vehicle Re-Identification (ReID) aims to retrieve all images of a given query vehicle from a large image database, without the license plate clues. The vehicles with different identities can be of the same make, model and color, while the vehicle appearances of the same identity always vary significantly across different viewpoints. Attention model can be applied for matching the key similarity of vehicle images across different viewpoints.

\vspace{5pt} \noindent \textbf{Datasets and Experimental Settings. } We conducted the experiments on two widely used vehicle datasets including Veri-776~\cite{veri776} and VehicleID~\cite{VehicleID}. Veri-776 dataset contains over 50,000 images from 776 vehicle IDs, where 37,778 images from 576 IDs are split for training and the rest 200 IDs are used for testing. VehicleID is composed of 110,178 images of 13,134 vehicles for training and 111,585 images of 13,133 IDs for testing. Following the experimental settings in~\cite{veri776} and ~\cite{lou2019embedding}, we report the testing results for three subsets including small size subset with 800 vehicles, medium size subset with 1,600 vehicles and large subset with 2,400 vehicles. We employed the CMC curve and mAP as the evaluation metrics for vehicle ReID task.

\vspace{5pt} \noindent \textbf{Implementation Details. }  We applied the ResNet50 backbone and the same attention block ($M\!\!=\!\!8$) as the baseline. The hyper-parameters are also fixed for the baseline and our method. The loss functions and data augmentation methods are same with person ReID task. We selected 256 samples in a batch with $256\times256$ image size. The initial learning rate is 2e-4 and reducing 10 times in 8000th and 18000th iterations. We trained the network for total 28000 iterations.

\vspace{5pt} \noindent \textbf{Results. } We compared the performance of CAL with the baseline attention learning method and other SOTA methods. As shown in Table~\ref{comparison_vehicle}, we obtained 0.9\%/2.3\% Rank-1/mAP improvement on the Veri-776 dataset and 5.8\%/3.3\%/4.1\% Rank-1 improvement on small/medium/large test settings of the VehicleID dataset.  Note that we did not use any extra labels in the training precess, yet achieved the comparable performance with VAML~\cite{Chu_2019_ICCV} which manually annotates the viewpoints of images to train the view-predictor.

\begin{table*}[t]
\caption{Comparisons with the state-of-the-art vehicle ReID methods on the VeRi-776 and VehicleID datasets.}
\label{comparison_vehicle}
\begin{center}
\vspace{-0.4cm}
\renewcommand\tabcolsep{4pt}
\begin{tabular}{l|*{3}{c}|*{3}{c}|*{3}{c}|*{3}{c}}
\hline
\multirow{3}{*}{\bf Method} &\multicolumn{3}{c|}{\bf Veri-776} &\multicolumn{9}{c}{\bf VehicleID} \\
\cline{2-13} &\multicolumn{3}{c|}{\bf Test 11587} &\multicolumn{3}{c|}{\bf Test 800} & \multicolumn{3}{c|}{\bf Test 1600} & \multicolumn{3}{c}{\bf Test 2400} \\
\cline{2-13} &{\bf R1} &{\bf R5} &{\bf mAP}& {\bf R1} &{\bf R5} &{\bf mAP}& {\bf R1} &{\bf R5} &{\bf mAP}&{\bf R1} &{\bf R5} &{\bf mAP}
\\ \hline
 GSTE ~\cite{bai2018group}  - &- & 59.4 & 87.1 &- &- &82.1 &- &-& 79.8& -& -\\
AAMI~\cite{zhou2018aware}  &85.9& 91.8& 61.3&   63.1 & 83.3& -&   52.9 &75.1 &-& 47.3 & 70.3 &- \\
FDA-NeT~\cite{Lou_2019_CVPR} & 84.3& 92.4& 55.5& -&-&-& 59.8 &77.1& 65.3& 55.5& 74.7 & 61.8\\
VAML$^*$~\cite{Chu_2019_ICCV}& 89.8 &96.0& 66.3 & 88.1 &97.3 &-& 83.2 &95.1 &-&  80.4& 93.0 &-\\
AAVER~\cite{Khorramshahi_2019_ICCV} & 88.7 & 94.1 & 58.5 &  72.5 & 93.2 & -&  66.9 & 89.4& -&  60.2& 84.9&-\\
EALN~\cite{lou2019embedding} &84.4 & 94.1&57.4 & 75.1 & 88.1 & 77.5 & 71.8 &83.9 & 74.2 & 69.3 & 81.4 & 71.0\\
DFLNet~\cite{Yan_2020_IJCAI} & 93.2 & 97.6 & 73.3 &  78.8 &\textbf{95.1} & 82.8 &-&-&-&  69.8 & \textbf{90.6} &75.4 \\
\hline
ResNet50  & 94.5 & 97.2 & 72.0 & 76.7 & 93.5 & 84.1 & 74.9 & 89.5 &81.4 & 71.0 & 84.9 &78.0 \\
ResNet50 + CAL & \textbf{95.4} & \textbf{97.9} & \textbf{74.3} & \textbf{82.5} & 94.7 & \textbf{87.8} & \textbf{78.2} & \textbf{91.0} & \textbf{83.8} & \textbf{75.1} & 88.5 & \textbf{80.9}\\
 \hline
\end{tabular}
\end{center}
\vspace{-0.7cm}
\end{table*}

\begin{table}[!]
\caption{Analysis of different counterfactual attentions. We implement different strategies to generate counterfactual attentions including random attention, uniform attention, reversed attention and shuffle attention. We report the top-1 classification accuracy of classification task. }
\label{tb:ca}
\begin{center}
\vspace{-0.3cm}
\renewcommand\tabcolsep{10.0pt}
\begin{tabular}{l|c|c|c}
\hline
 & \textbf{CUB} & \textbf{Cars} & \textbf{Aircraft} \\
\hline
Baseline & 89.3 & 94.0 & 93.6 \\
\hline
Random Attention & \textbf{90.6} & \textbf{95.5} &94.2 \\
Uniform Attention & 90.2 & 95.3 &  94.2 \\
Reversed Attention & 89.2 & 94.1 & 93.2 \\
Shuffle Attention & 90.4  & 94.3  &  \textbf{94.5}\\
 \hline
\end{tabular}
\end{center}
\vspace{-0.6cm}
\end{table}

\begin{figure}[!t]
  \centering
  \includegraphics[width=\linewidth]{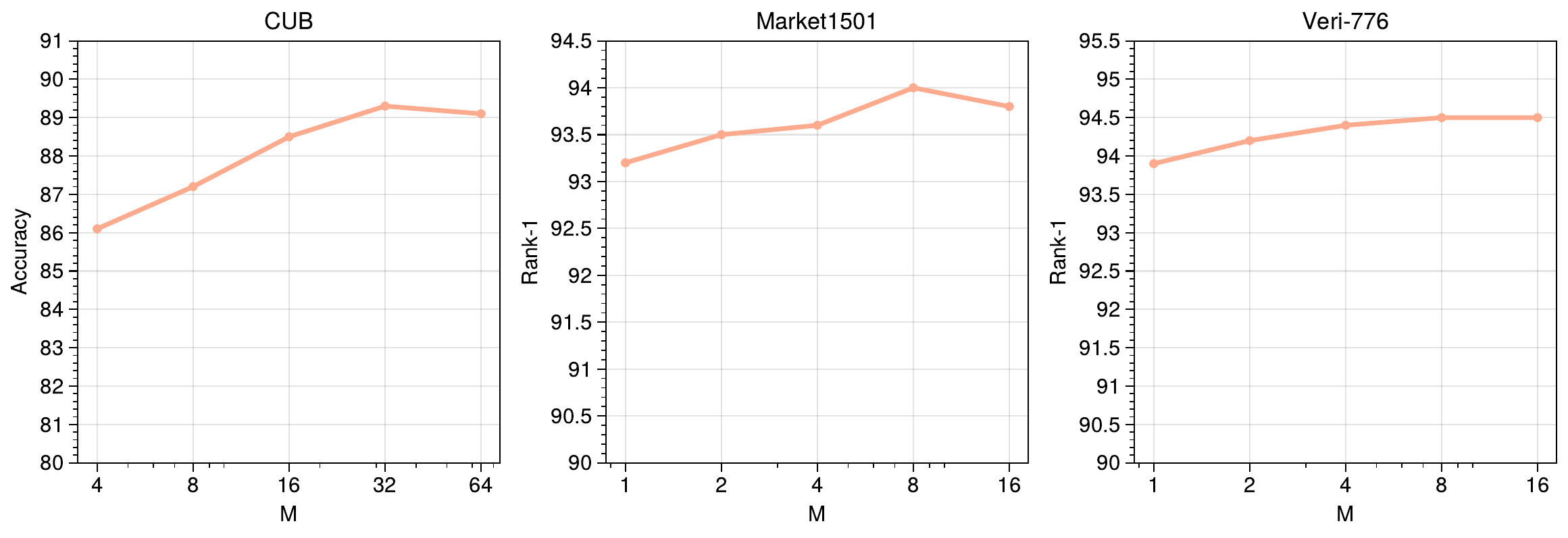}
   \vspace{-10pt}
  \caption{ Effects of the number of attentions. We investigate the effects of the numbers of attention on CUB, Market1501 and Veri-776 for fine-grained image categorization, person re-identification and vehicle re-identification respectively and directly use the best hyper-parameters on other datasets.}
\vspace{-0.6cm}
\label{fig:att}
\end{figure}

\begin{table}[!]
\caption{ Quantitative analysis of attention. We compare the classification accuracy (\%) our method with other three kinds of attention regularization strategies including attention drop, entropy regularization and attention normalization,  and evaluate the quality of the learned attention maps using mIoU  (\%)  with ground-truth bounding boxes on CUB. }
\label{tb:iou}
\begin{center}
\vspace{-0.3cm}
\renewcommand\tabcolsep{10.0pt}
\begin{tabular}{l|c|c}
\hline
 & \textbf{Accuracy} & \textbf{mIoU} \\
\hline
Baseline &  89.3 & 54.2  \\
\hline
+ Attention Dropout & 88.9 & 50.3 \\
+ Entropy Regularization & 88.7 &  51.1\\
+ Attention Normalization & 85.8  & 46.2\\
\hline
+ CAL &  \textbf{90.6} &  \textbf{67.4}  \\
 \hline
\end{tabular}
\end{center}
\vspace{-0.6cm}
\end{table}

\begin{table}[!t] \small
\caption{Results of single-head attention models.  We report the top-1 classification accuracy (\%) on three fine-grained categorization datasets. }
\label{tb1}
\begin{center}
\vspace{-0.4cm}
\renewcommand\tabcolsep{10.0pt}
\begin{tabular}{l|c|c|c}
\hline
 & \textbf{CUB} & \textbf{Cars} & \textbf{Aircraft} \\
\hline
Baseline ($M\!\!=\!\!1$)  & 85.9 & 92.1 & 91.5 \\
 + CAL & \textbf{88.2} & \textbf{94.2} & \textbf{92.9} \\
 \hline
\end{tabular}
\end{center}
\vspace{-0.6cm}
\end{table}

\subsection{Analysis} \label{sec:ablation}

We analyzed the influences and sensitivity of some major parameters. We conducted the parameters analysis
experiments on three fine-grained visual recognition tasks.

\vspace{5pt} \noindent \textbf{Effects of the type of counterfactual attention. } We investigated three different strategies to generate the counterfactual attention maps, namely random attention, uniform attention, reversed attention and shuffle attention  (see Supplementary Material for details). The results are presented in Table~\ref{tb:ca}. We see random attention, uniform attention and shuffle attention achieve similar performance while reversed attention fails to improve the baseline on CUB. We think it is because learning attention that is better than reversed attention is relatively easy and cannot provide an effective signal to supervise the attention. 

\vspace{5pt} \noindent \textbf{Effects of the number of attentions.} The number of heads in attention model is an important hyper-parameter in our baseline model. Therefore, we search the best numbers of attention on CUB~\cite{cub}, Market1501~\cite{market1501} and Veri-776~\cite{veri776} for fine-grained image categorization, person re-identification and vehicle re-identification tasks respectively and directly use the searched hyper-parameters on other datasets. For a fair comparison, we use the same hyper-parameters in our models and the baseline models, and did not search the best hyper-parameters for our models separately. The results are presented in Figure~\ref{fig:att}. Based on these results, we set $M$ to 32 and 8 for fine-grained categorization and re-identification tasks, respectively.

\begin{figure*}[t]
  \centering 
  \includegraphics[width=0.94\linewidth]{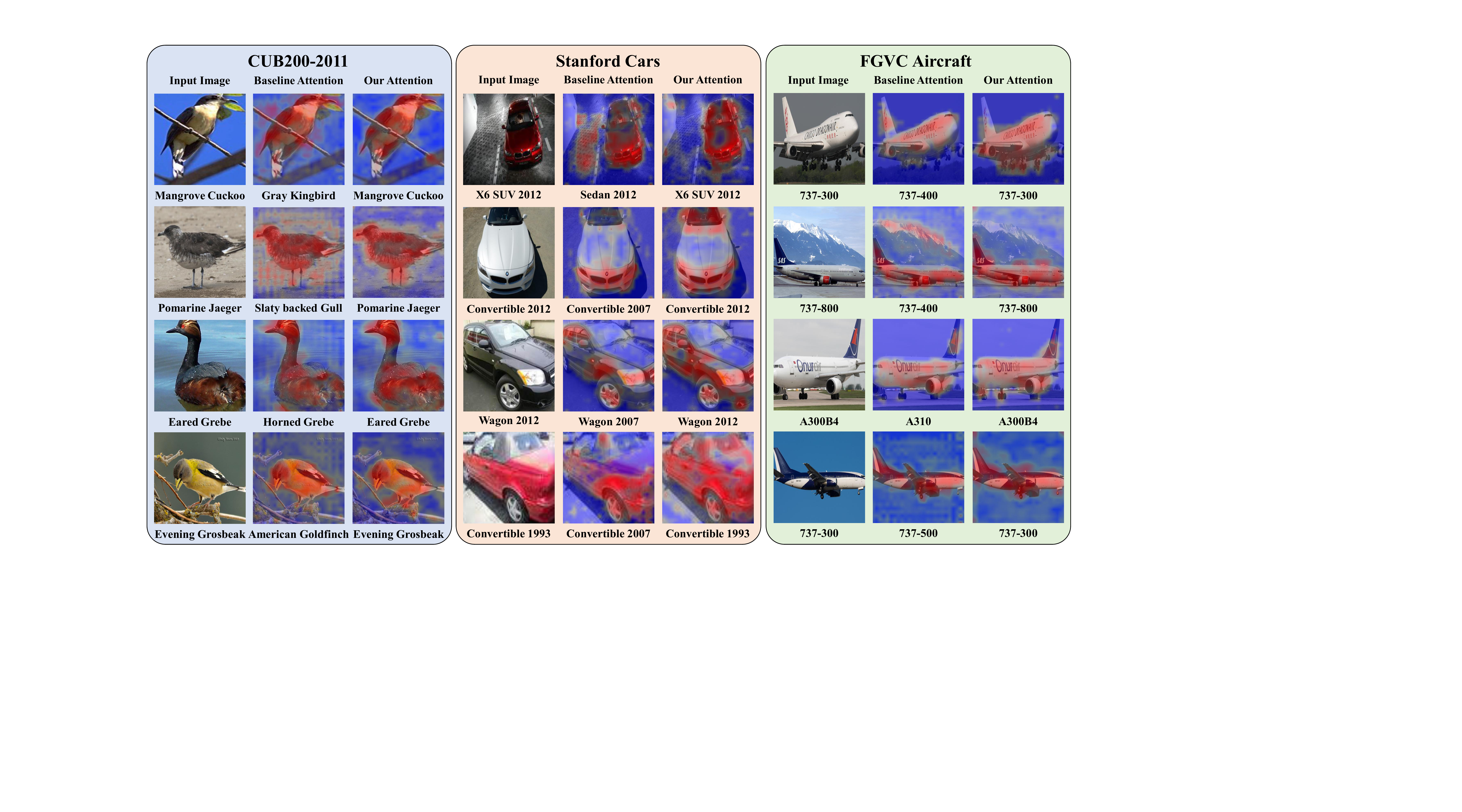}
  \caption{Visualization of the attention maps of our models and the baseline models. We see our method helps the attention models make correct predictions by 1) reducing the misleading and scatter attentions and 2) encouraging the model to focus on the main clues for classification and explore more discriminative regions.    Best in color.
}
\label{fig:compare2}
\end{figure*}

\vspace{5pt} \noindent \textbf{Quantitative analysis of attention.} To better verify the effectiveness of CAL, we compare our method with other three kinds of attention regularization strategies including attention drop, entropy regularization and attention normalization (see Supplementary Material for details) and evaluated the quality of the learned attention maps by computing the mean IoU between the rectangular
region that covers the high score attentions and the ground-truth object bounding boxes on CUB. The results can be found in Table~\ref{tb:iou}. We see only CAL is effective to simultaneously improve classification accuracy and attention quantitative. Both attention Dropout and Entropy regularization will slightly degrade the final performance under the both metrics. Attention normalization will significantly hurt the performance. 

\vspace{5pt} \noindent \textbf{Results of single-head attention models. } To show the generality for different attention models, we also test CAL on single-head attention models (\emph{i.e.}, baseline models with $M\!\!=\!\!1$). The results are presented in Table~\ref{tb1}. We see our method can consistently and more significantly improve the relative weak baseline models, which clearly shows our method is suitable for various attention models.

\subsection{Visualization}

To have an intuitive understanding of our counterfactual attention learning method, we compare the attention maps of our models and the baselines models on CUB200-2011~\cite{cub}, Stanford Cars~\cite{cars}  and FGVC-Aircraft datasets~\cite{aircraft}. The visual results are displayed in Figure~\ref{fig:compare2}. We see our method helps the attention models make correct predictions by reducing the misleading and scatter attentions. For example, in the first example of the Stanford Cars dataset, the attention with our CAL method avoids the reflection on the ground. Besides, CAL encourages the model to focus on the main clues for classification and explore more discriminative regions. Taking the Eared Grebe in the CUB200-2011 dataset as example, our attention focuses on the discriminative buttocks region to recognize it. While for the second example of cars and the first one of aircrafts, our attention models tend to explore more discriminative regions such as the rearview mirror and wheel respectively.

\section{Conclusion}

In this paper, we have presented a counterfactual attention learning method to learn more effective attention based on causal inference. We designed a framework to quantitate the quality of attentions by comparing the effects of facts and the counterfactuals on the final prediction. We also proposed to maximize the difference to encourage the network to learn more effective visual attentions. Our method only brings negligible extra cost during training and introduce no cost during inference. 
CAL is a model-agnostic framework to enhance attention learning and mitigate the effects of dataset bias, which can be applied to various fine-grained visual recognition tasks. 
We conducted extensive experiments on three fine-grained visual recognition tasks and demonstrated state-of-the-art performance on all benchmarks.

\section*{Acknowledgements}
This work was supported in part by the National Key Research and Development Program of China under Grant 2018AAA0102803, in part by the National Natural Science Foundation of China under Grant 61822603, Grant U1813218, and Grant U1713214, in part by a grant from the Beijing Academy of Artificial Intelligence (BAAI), and in part by a grant from the Institute for Guo Qiang, Tsinghua University.

\section*{A. More Visual Results}

\begin{figure*}
  \centering
  \includegraphics[width=0.65\linewidth]{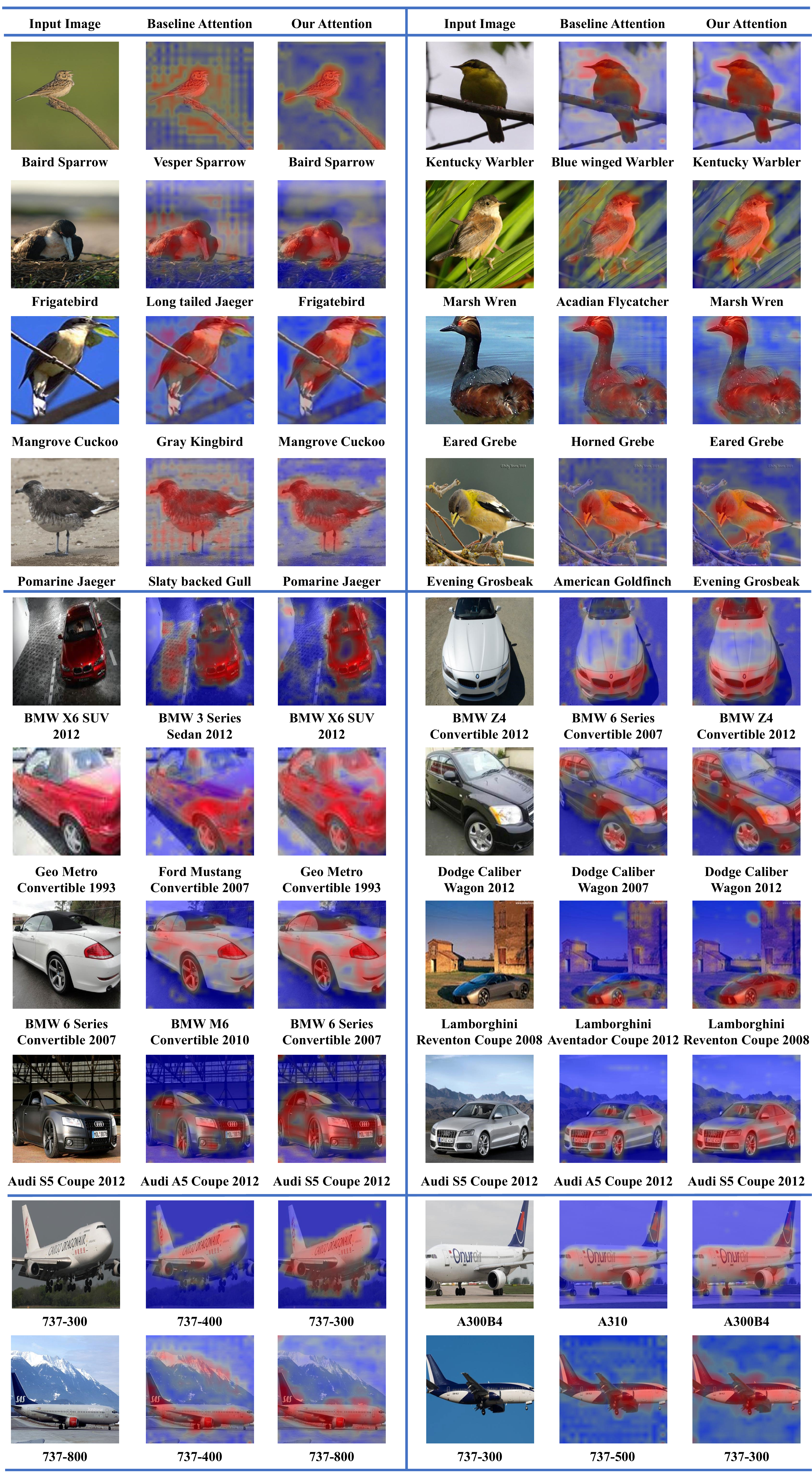}
  \caption{Visualization of the attention maps of our models and the baseline models. We see our method helps the attention models make correct predictions by 1) reducing the misleading and scatter attentions and 2) encouraging the model to focus on the main clues for classification and explore more discriminative regions. Best viewed in color.
}
\label{vis}
\vspace{-0.3cm}
\label{fig:compare}
\end{figure*}

To have an intuitive understanding of our counterfactual attention learning method, we compare the attention maps of our models and the baselines models on CUB~\cite{cub}, Stanford Cars~\cite{cars}  and Aircraft datasets~\cite{aircraft}. The more visual results are presented in Figure~\ref{vis}. We see our method helps the attention models make correct predictions by 1) reducing the misleading and scatter attentions and 2) encouraging the model to focus on the main clues for classification and explore more discriminative regions.

\section*{B. More Implementation Details}

\paragrapha{5pt}{Different types of counterfactual attentions. } We compared four different counterfactual attentions in our experiments. The details about how to generate them are described as follows. 

\begin{itemize}
    \item \textbf{Random Attention. } We use randomly generated attention maps as the counterfactual attentions. The attention value for each location is sampled from a uniform distribution $\mathcal{U}(0,2)$.
    \item \textbf{Uniform Attention. } We simply set the attention value for each location to the average value of the real attention maps.
    \item \textbf{Reversed Attention. } We reverse the attention maps by subtracting the original attention from the maximal attention value of each sample.
    \item \textbf{Shuffle Attention. } We randomly shuffle the attention maps along the batch dimension.
\end{itemize}

\paragrapha{5pt}{Attention Regularization Strategy. } We investigated several regularization strategies on the baseline attention model to verify the effectiveness of our method. The details about these  regularization strategies are described as follows. 

\begin{itemize}
    \item \textbf{Attention Dropout. } We apply the Dropout method to the attention maps.
    \item \textbf{Entropy Regularization. } We add an extra term to the loss function to maximize the entropy of the attention maps.
    \item \textbf{Attention Normalization. } We add $\ell_2$ normalization to the attention maps.
\end{itemize}

{\small
\bibliographystyle{ieee_fullname}
\bibliography{reference}
}

\end{document}